\documentclass{article}
\usepackage{ijcai23}

\usepackage{times}
\usepackage{soul}
\usepackage{url}
\usepackage[hidelinks]{hyperref}
\usepackage[utf8]{inputenc}
\usepackage[small]{caption}
\usepackage{graphicx}
\usepackage{amsmath}
\usepackage{amsthm}
\usepackage{booktabs}
\usepackage{algorithm}
\usepackage{algorithmic}
\usepackage[switch]{lineno}
\usepackage{amssymb}
\usepackage{xcolor}
\usepackage{multirow}

\definecolor{gjc}{rgb}{0, 0.5, 0}

\title{WeatherFormer: Empowering Global Numerical Weather Forecasting with Space-Time Transformer}

\author{
Junchao Gong$^{1,2}$
\and
Tao Han$^2$\and
Kang Chen$^2$\and
Lei Bai$^2$
\affiliations
$^1$Shanghai Jiao Tong University\\
$^2$Shanghai AI Laboratory
\emails
baisanshi@gmail.com
}

\begin{document}

\maketitle

\begin{abstract}
% Accurately predicting weather state guides the social production scientifically and ensures the safety of life and property for human. However, traditional Numerical Weather Prediction (NWP) system resolves it by solving complex partial differential equations with high-performance computing clusters, resulting in tons of carbon emission. Recently, Artificial Intelligence (AI) based weather predictors have been proposed to chase the physical method in a more ecological way. To further narrow the performance gap between the AI-based methods and physic heuristic predictor, this paper proposes WeatherFormer, a new transformer-based NWP framework, to model the complex spatio-temporal atmosphere dynamics and empower the capability of data-driven NWP. Firstly, WeatherFormer introduces space-time factorized transformer blocks to decrease the model parameters and memory consumption. Also, two data augmentation strategies customized to the earth data are introduced to boost the performance. 
% Extensive experiments on benchmarking dataset show WeatherFormer achieves superior performance against existing deep learning methods and further narrows the gap to approach advanced physical models.

Numerical Weather Prediction (NWP) system is an infrastructure that 
%has a great influence 
exerts considerable impacts
on modern society.
% Traditional NWP system relies on solving complex partial differential equations, 
% % built upon physical mechanisms, 
% which is extremely expensive in computation and energy. 
Traditional NWP system, however, resolves it by solving complex partial differential equations with a huge computing cluster, resulting in tons of carbon emission. 
% With the matureness of deep learning techniques and the accumulation of large scale weather datasets, 
Exploring efficient and eco-friendly solutions for NWP attracts 
 interests from Artificial Intelligence (AI) and earth science communities. 
 To narrow the performance gap between the AI-based methods and physic predictor,
% However, there is still a considerable gap between physical and data-driven methods in terms of prediction performance.
this work proposes a new transformer-based NWP framework, termed as WeatherFormer, to model the complex spatio-temporal atmosphere dynamics and empowering the capability of data-driven NWP. WeatherFormer innovatively introduces the space-time factorized transformer blocks to decrease the parameters and memory consumption, 
% in modeling global atmosphere data
in which Position-aware Adaptive Fourier Neural Operator (PAFNO)
% is proposed for efficient token mixing.  
is proposed for location sensible token mixing. 
Besides, two data augmentation strategies 
% ,motivated by the longitudinal rotation symmetry and discreteness in the data, 
are utilized to boost the performance and decrease training consumption.
Extensive experiments on WeatherBench dataset show WeatherFormer achieves superior performance over existing deep learning methods and further approaches the most advanced physical model.

% Extensive experiments on WeatherBench dataset show WeatherFormer achieves superior performance over existing deep learning methods and further approaches the most advanced physical model.

% We hope our work can motivate more future works to explore the potential of data-driven methods for NWP.

% Additional experiments on Moving MNIST and TrafficBJ also demonstrate the potential of WeatherFormer in diverse scenarios. 

\end{abstract}

\section{Introduction}
\begin{figure}[t]
  \centering
      \includegraphics[width=0.5\textwidth]{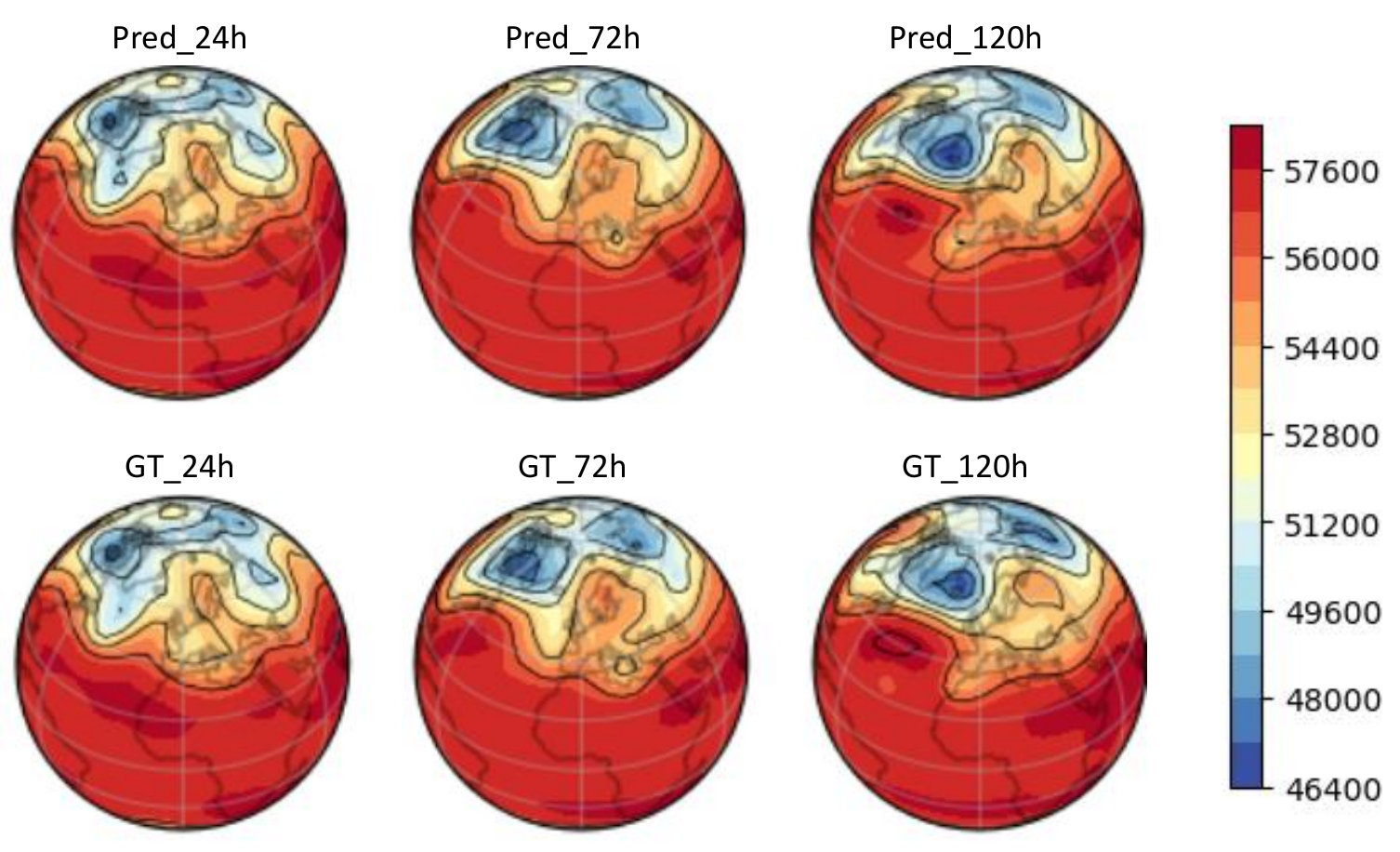}
    \caption{The prediction (Pred) and ground truth (GT) geopotential on the height of 500 hPa at 24-hour, 72-hour, and 120-hour, separately.}
    \label{fig:z500_pred}
    \vspace{-2mm}
\end{figure}

% As crucial infrastructure, numerical weather prediction (NWP) aims to predict future weather states (e.g., temperature, wind speed, humidity, etc.),  which significantly impacts our society.  Accurate NWP supports producing activities such as crop yields and energy production.
% % , and 
% Furthermore,  the extreme weather prediction can save losses from high-impact weather like floods and forest fires, which cause thousands of death and billions of dollars in economic losses every year.

Weather prediction plays a decisive role in various productive activities, as well as extreme weather events' prevention~\cite{reichstein2019deep}. For instance, accurate weather prediction can provide fundamental information for agricultural planting and irrigation system management. Also, it can help prevent massive life and property loss by forecasting typhoons, heatwaves, tropical cyclones, floods, etc. In particular, a series of floods that occurred in South East Queensland, the Wide Bay–Burnett, and parts of coastal New South Wales in 2022, has been one of the nation's recorded flood disasters, where 22 people are known to have died during the disaster. Many miserable natural events warn us that it is of great value to develop numerical weather prediction (NWP) to predict future weather states (e.g., temperature, wind speed, humidity, etc.), because it is the groundwork of the weather prediction and significantly impacts our society.

Modern NWP utilizes physics and fluid mechanics to model the atmosphere with complex partial differential equations (PDE)~\cite{bauer2015quiet}. 
To obtain solutions of the PDEs, initial states are first derived by processing data collected from satellites and other sensors. Then the solutions of discretized governing equations and parameterized sub-grid processes~\cite{kalnay2003atmospheric} can be obtained by using numerical techniques. However, due to the huge computational cost of the fine grids for high-frequency wave modeling and model ensemble in probabilistic forecasts, obtaining solutions in physics based models is time costly and intensively energy consuming. For example, it takes 82 minutes for the high-resolution Integrated Forecasting System (IFS) model to compute a 15-day, 51-member ensembled weather state prediction by using the ``L91'' 18km resolution grid data on the 1530 Cray XC40 nodes with dual socket Intel Haswell processors~\cite{bauer2020ecmwf}. Power consumption of Cray XC40 is 1900KW ~\cite{TOP500}, which dramatically higher than a single A100 GPU whose power is 400W.

Recently, the increasing interest raised in the deep neural networks based data-driven NWP methods stems from the following aspects:
% {\color{gjc} (1) save energy and give detail as rebuttal.}
(1) Data-driven NWP is consistent with the  Sustainable Development Goal in sustainable infrastructure.
% The power of Cray XC40 is 1800KW, while a single A100 GPU holds a power of 400W and  possesses the potential to compare with  Cray XC40 in NWP ~\cite{pathak2022fourcastnet} ~\cite{ECMWF}.
(2) As the increasing amount of available weather state data, the physics based NWP methods either fall short in the ability to incorporate signals from newly emerging geophysical observation systems~\cite{goodman2019goes}, or cannot efficiently process the Petabytes-scale NWP observation data; (3) Data-driven NWP methods enables large scale of ensembles with relatively low computational cost for the probabilistic forecasts and data assimilation~\cite{pathak2022fourcastnet}. 
% (3) Data-driven models can avoid biases in physics models such as those in convection parameterization schemes that strongly affect precipitation forecasts~\cite{balaji2021climbing}~\cite{schultz2021can}. 
Previous works~\cite{rasp2020weatherbench,weyn2020improving,rasp2021data,pathak2022fourcastnet,chattopadhyay2022towards} attempted to accurately predict future weather states in an efficient way by using data-driven methods. For example, FourCastNet~\cite{pathak2022fourcastnet} can predict high-resolution weather states with a reduction on the computational cost by a factor of 1000 when compared with IFS. However, there is still a performance gap between the physics based and data-driven NWP methods.

It is observed that the change of weather states heavily depends on their contextual weather states. 
% It is observed that different locations on Earth come with various terrains, which could easily affect the change of weather states.
Also, intuitively, the trend on the change of the weather states over the past long period of time would have a great impact on the future weather states. Therefore, in addition to only considering the spatial information at the current time step, the temporal relationship over the past long period of time should be well modeled. However, the existing deep neural networks based data-driven numerical weather prediction methods~\cite{pathak2022fourcastnet,rasp2020weatherbench,weyn2020improving,rasp2021data} cannot process spatio-temporal information without expensive multisteps finetuning.

Based on the above observations, we propose a new deep neural networks based numerical weather prediction framework called WeatherFormer, which takes the weather states at several time steps over the past to model the spatio-temporal information simultaneously and produce future weather states for a long period of time. Our WeatherFormer is a transformer based deep neural network, which is composed of a set of space-time factorized blocks (SF-Block). Specifically, within each SF-Block, a spatial mixer and a temporal mixer are used to mix the spatio-temporal information. Note that both the spatial mixer and the temporal mixer have the same structure but one operates on the spatial domain and the other on the temporal domain. In order to reduce the computational cost, the single filter strategy proposed by adaptive Fourier neural operator (AFNO)~\cite{guibas2021adaptive},  is partly adopted
% the single filter strategy 
% proposed by adaptive Fourier neural operator (AFNO)~\cite{guibas2021adaptive} 
within each mixer, where the input is first transformed into frequency domain to mix information and then is reversed back into the original domain. Additionally, based on the AFNO, we introduce a novel position-aware adaptive Fourier neural operator (PAFNO) to encode relative position information in the spatial domain by assigning different coefficients to different frequency filters during the information mixing process. Moreover,  earth rotation augmentation is applied to exploit  rotation equivariance and noise augmentation to obtain a comparable multi-step performance with half of training consumption.

% {\color{gjc} paragraph about data augmentation}Moreover, it is well-known that deep neural networks can easily suffer from the overfitting problem. We then find that the weather states horizontally rotate as the Earth rotates over the axis of the Earth, which brings rotation equivariance to the weather state data. Based on this observation, the Earth Rotation augmentation is introduced to alleviate overfitting. 
% Also, it is observed that for a long-term prediction, error accumulation often happens when using an autoregressive strategy, which can easily lead to performance degradation. To this end, we introduce Error Overlapping augmentation, where random noise is added during training to enhance the long-term robustness of our WeatherFormer.

The contributions of this work are summarized as follows:
\begin{itemize}%[noitemsep, topsep=0pt]
    \item We propose WeatherFormer, a new data-driven numerical weather prediction framework based on a spatio-temporal Transformer, which can predict future weather states by considering spatio-temporal information of the past weather states.
    \item We introduce the Position-aware Adaptive Fourier Neural Operator (PAFNO), which could capture position information of weather signals while maintaining low parameters and computation cost. % with Fourier neural operators.
    \item 
    % {\color{gjc} contribution about augmentation}
    We introduce the Earth Rotation augmentation to take advantage of the rotation equivariance of the data to ease the overfitting issue.
    % and the Error Overlapping augmentation to relieve the negative impact of the error accumulation problem in the autoregressive long-term prediction strategy.
    \item Extensive experiments on the WeatherBench dataset demonstrate the effectiveness of our proposed WeatherFormer over strong data-driven NWP methods.
\end{itemize}

\section{Related Works}
\subsection{Traditional Numerical Weather Prediction}
% 1.1
Traditional NWP can trace back to the 20th century. Bjerknes and Abbe recognized that predicting the state of the atmosphere could be treated as an initial value problem of mathematical physics~\cite{bjerknes1904problem}, wherein future weather is determined by integrating the governing partial differential equations which starting from the observed current weather~\cite{bauer2015quiet}.
% 1.2 how to solve
Researchers numerically solved prognostic equations build upon Navier–Stokes equations, the ideal gas law, and other physical equations which are intractable to obtain analytical solutions~\cite{kalnay2003atmospheric}. 
At first, the initial state (called the analysis) of the atmosphere and surface is derived as a Bayesian inversion problem using observations, prior information from short-range forecasts and their uncertainties as constraints as well as the forecast model~\cite{lorenc1986analysis}~\cite{daley1993atmospheric}. 
Then, numerical techniques such as spectral methods and finite-difference methods are chosen according to numerical stability, accuracy, and computational speed  to attain solutions~\cite{robert1982semi}.
%1.3 NWP的挑战和问题
However, physic-based methods are computationally expensive for model ensemble and solving PDEs with fine grids. The ECMWF 16-km highest-resolution model, which performs calculations on two million grid columns with 10-min time stepping over a 10-day period, consumes about 4 MVA power for each prediction~\cite{bauer2015quiet}.
\subsection{Data-driven Numerical Weather Prediction}
% 2.1
Data-driven NWP models predict future weather by extracting statistical regularities from historical weather data.
% 2.2
%数据集
Rasp introduced the WeatherBench dataset as a benchmark challenge for data-driven medium-range weather~\cite{rasp2020weatherbench}.
%backbone
% Motivated by ~\cite{rasp2020weatherbench}, researchers applied classical deep learning models, such as ResNet and UNet ~\cite{he2016deep} ~\cite{ronneberger2015u}, for NWP ~\cite{rasp2021data} ~\cite{weyn2020improving}.
Motivated by~\cite{rasp2020weatherbench}, classical deep learning models such as ResNet~\cite{he2016deep} and UNet~\cite{ronneberger2015u} are applied for NWP.
Compared with naive CNN~\cite{lecun1995convolutional}, UNet and ResNet remarkably enhance prediction results and even outperform physical model T64~\cite{rasp2020weatherbench} as shown in Table~\ref{tab:nwp baselines}. Our work explore the potential of the transformer, which achieves conspicuous success in the computer vision research community~\cite{vaswani2017attention} . 
% 数据增强
To further improve predictions, researchers train models with additional data.
Rasp applied additional simulated data from Coupled Model Intercomparison Project (CMIP) for pretraining~\cite{rasp2021data}. To increase data for training, Chattopadhyay adopts spatial transformer network (STN)~\cite{jaderberg2015spatial} to simulate rotation and translation on atmosphere~\cite{chattopadhyay2022towards}. Although the above methods attain promising results, the curiosity leads us to explore a naive method for generating additional data.
% we are curious about a naive way to generate additional data.
% 减少参数
% For high-resolution weather forecasting, 
Pathak proposed FourCastNet which applies AFNO to save parameters while decreasing computation complexity~\cite{pathak2022fourcastnet}. However, as a trade-off, AFNO cannot introduce position relations between tokens to the model. To attain position information, we designed a new Fourier neural operator which only introduces $N$ (the length of tokens) additional parameters.

\section{Methodology}
\label{method}
\begin{figure}
  \centering      \includegraphics[width=0.47\textwidth]{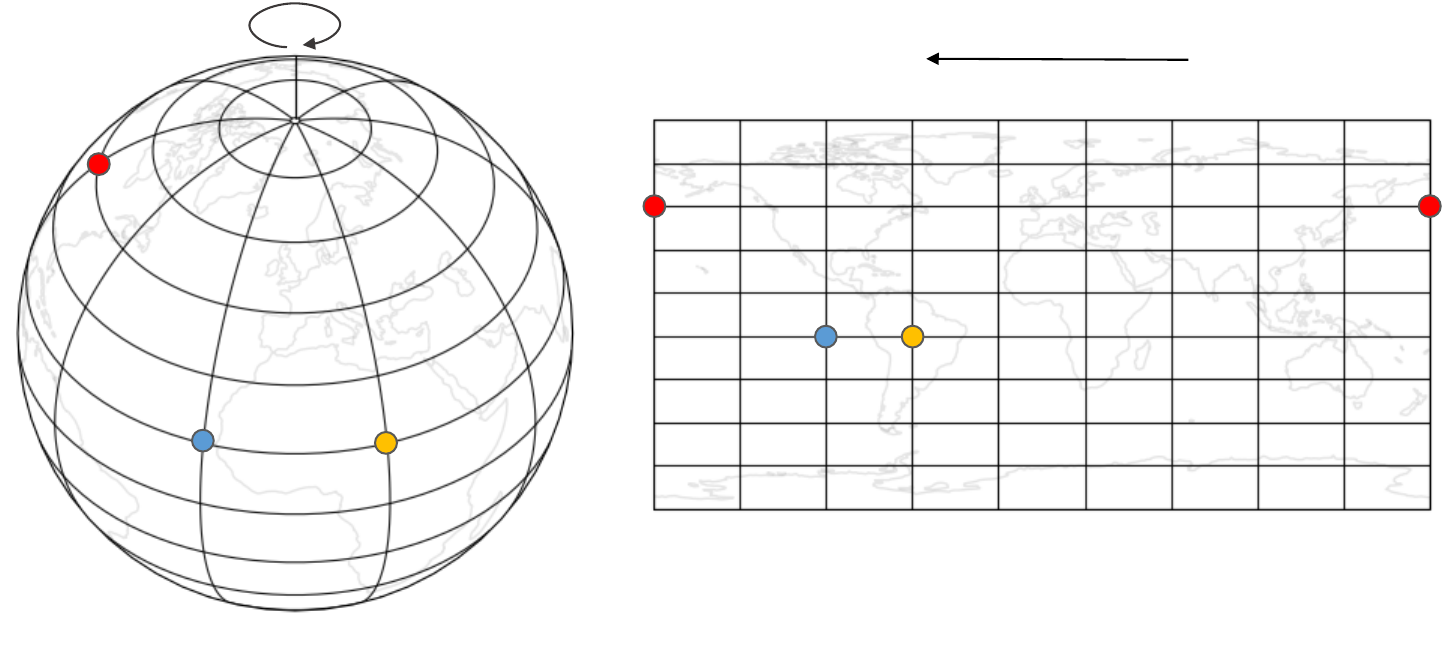}
    \caption{Mapping data points with the same color from sphere to plate. Note that, the red point is mapped to the left and right sides of the plate which means the left and right sides are continuous. Arrows indicate rotation in the sphere and shifting in the plate. }
    \label{fig:point_mapping}
\end{figure}

\begin{figure*}[t]
  \centering
    \scalebox{1}{
      \includegraphics[width=1\textwidth]{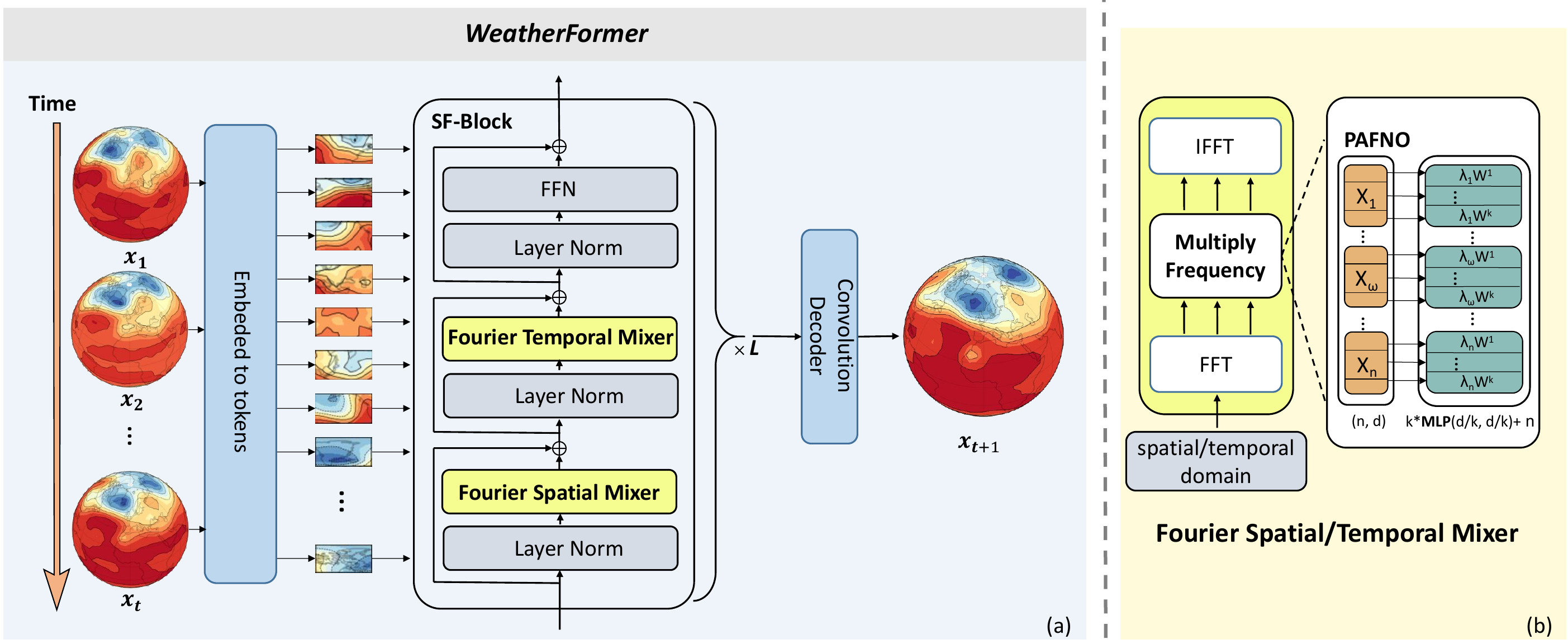}
      }
    \caption{ (a) Overview of WeatherFormer. It first divides a sequence of weather states into patch tokens. These tokens are then processed by $L$ layers of SF-Block, which contains a Fourier spatial mixer and a Fourier temporal mixer. Finally, a convolution decoder decodes the output tokens to the future weather states. (b) Details of our Fourier mixer. It first transforms token features to frequency domain with fast Fourier transform. Then, frequency features are multiplied by PAFNO filters which consist of $k$ Multilayer Perceptrons (MLPs) and $n$ frequency coefficients. Finally, frequency tokens are  transformed back to spatial/temporal domain. 
    % \bai{Fourier Spatial Mixer; the figure can be divided into two figs so that you can make the left one more beautiful}
}
    \label{fig:framework_overview}
    % \vspace{-2mm}
\end{figure*}

\subsection{Framework Overview} \label{sec:overview}
Our WeatherFormer takes the weather states of previous time steps as input and outputs predicted weather states of future time steps. For each time step, the weather states are sampled at $H \times W$ locations along the latitude axis and the longitude axis from the whole Earth (shown in Figure~\ref{fig:point_mapping}), where $H$ and $W$ are the number of sampled locations along the latitude axis and the longitude axis, respectively. We denote the input data as $\mX\in \Rbb^{T\times H\times W\times C}$, where $T$ is the number of previous time steps used as input, and $C$ is the number of weather states (e.g., temperature, wind speed, humidity, etc.) at each sampled location. For each forward pass, our WeatherFormer predicts the weather states of the next time step $\mY \in \Rbb^{H\times W\times C}$, and it can autoregressively generate multistep predictions. The formulation of the autoregressive prediction is represented below:
\begin{align}
    \hat{\mY}_{T+1} &=WeatherFormer(\mX_{1:T})
    \\
    \hat{\mY}_{T+2} &=WeatherFormer([\mX_{2:T}, \hat{\mY}_{T+1}])\\
    \vdots
    \\
    \hat{\mY}_{T+n} &=WeatherFormer([\hat{\mY}_{n}, \dots, \hat{\mY}_{T+n-1}])
    \label{eq:question_formulation}
\end{align}

As shown in Figure \ref{fig:framework_overview}, our WeatherFormer is built upon transformer~\cite{vaswani2017attention}, where the input is split and embedded into tokens. Then these tokens are entered into a series of transformer blocks to pass messages among tokens. Instead of using the traditional transformer blocks,  factorized space-time Block (SF-Block) is designed with a Position-aware Adaptive Fourier Neural Operator (PAFNO) in it to effectively model the spatio-temporal information within the past weather states and encode their positional information. After that, a convolution decoder is applied to recover the encoded tokens into future weather states.

% {\color{gjc}Paragraph about data augmentation}
Additionally,  two augmentation strategies are implemented:1)  earth rotation augmentation that leverages the rotational equivariance property of the data to ease the overfitting issue of our proposed WeatherFormer; and 2) noise augmentation, which decrease the prediction  error accumulation caused by the autoregressive strategy.

\subsection{Factorized Space-time Block}
Existing data-driven NWP methods~\cite{pathak2022fourcastnet}~\cite{rasp2021data}~\cite{rasp2020weatherbench}~\cite{chattopadhyay2022towards} predict future weather states by only considering the spatial information over the current states. However, the change of the weather states is dynamic over time, and there is abundant information about the trend of the weather states within the historical weather states data, which could benefit the prediction. Therefore, instead of using only the current states, it is reasonable to use a set of weather states at previous consecutive time steps to predict the weather states at the next time step, and an elegant design is needed to explore the spatio-temporal information over the input data. To this end, inspired by the work in~\cite{arnab2021vivit}, we 
adopt % propose 
an SF-Block to mix the information over the previous weather states in both spatial and temporal domains.

As shown in Figure~\ref{fig:framework_overview}, our SF-Block factorizes the self-attention module in a traditional transformer block by introducing a spatial mixer and a temporal mixer to separately model the spatial and temporal relationship over the input tokens.
Given ${\mZ} \in \Rbb^{B\times t\times h\times w\times C}$ as the input of our SF-Block, $\mZ$ is reshaped to $\mZ_{s} \in \Rbb^{(B \cdot t)\times h\times w\times C}$ at first.
The Fourier Spatial Mixer calculates spatial attention for $\mZ_s$.
Tokens obtained after the Fourier Spatial Mixer, $\mZ_s^\prime \in \Rbb^{(B \cdot t)\times h\times w\times C}$, are reshaped to $\mZ_{t} \in \Rbb^{(B \cdot h \cdot w)\times t\times  C}$.
The  Temporal Mixer calculates the temporal attention along the temporal axis $t$, which implies that the batch size of $\mZ_{t} $ is $B \cdot h \cdot w$ and the length of the token sequence is $t$.
In this way, the computational cost can be significantly reduced.

% Specifically, we denote the input tokens to the $l$-th SF-Block in our WeatherFormer as $\mZ^{l} \in \Rbb^{\frac{T}{t}\times \frac{H}{h} \times \frac{W}{w} \times D}$, where $t$, $h$, $w$ are the patch size used to tokenize the input $\mX$, and $D$ is the feature dimension of the tokens.

Note that PAFNO is adopted for both the spatial mixer and the temporal mixer, which is different from ViVIT~\cite{arnab2021vivit} for computing attention in the frequency domain.
% in both the spatial mixer and the temporal mixer, 
% and the only difference is that one mixes information only over the spatial domain while the other over the temporal domain.
Details of the PAFNO are in the following section.

\subsection{Position Aware Adaptive Fourier Neural Operator}

Our PAFNO is built upon AFNO~\cite{guibas2021adaptive}. In AFNO, the input tokens are first transformed into the frequency domain by applying a discrete Fourier transform operation. After that, each element in the transformed token sequence is fed into a two-layer MLP to mix information. In order to reduce the computational cost, instead of assigning different MLP weights for each element in the transformed token sequence, the weights of the MLP layers are shared across the whole transformed token sequence. Finally, inverse discrete Fourier transform operation is applied to reverse the processed token sequence back into the original domain. 

For the NWP task, at a specific location, the previous weather states at its neighboring position have more impact on the future weather state prediction than those at distance positions~\cite{kalnay2003atmospheric}. Therefore, the position information is critical to the NWP task. However, the original positional embedding strategy used in~\cite{guibas2021adaptive} does not work well. 
% One possible explanation is that the position embedding information disappears when the input tokens are transformed into the frequency domain.  
Moreover, as the MLP weights used in AFNO are shared among tokens at all positions, each position is considered equally important, which neglects the impact of positional information displayed as Figure~\ref{fig:weights}. To this end, we employ a set of learnable position-related coefficients $\{\lambda_{n}\}^N_{n=1}$ to assign a coefficient for the token at each position, where $N$ is the number of tokens. Thorough analytics about position embedding, its failure in frequency mixers, and the way PAFNO introduces position information are provided below.

\vspace{-12pt}
\subsubsection{Position Embedding}
In the self-attention mechanism of the traditional transformer, the absolute position embedding introduces the positional information to the attention weight matrix. Given the input $\mX$ and the corresponding position embedding $\mP$, the $i$-th output mixed token $\vo_i$ of the self-attention is formulated as follows:
\begin{align}
\mQ &=\left(\mX+\mP\right) \boldsymbol{W}_Q , \label{eq:Q}\\ 
\mK &=\left(\mX+\mP\right) \boldsymbol{W}_K , \label{eq:K}\\ 
\mV &=\left(\mX+\mP\right) \boldsymbol{W}_V , \label{eq:V}\\ 
a_{i, j} &=\frac{\exp(\vq_i \vk_j^{\top})}{\exp(\sqrt{d} \cdot \sum_{m} \vq_i \vk_m^{\top})} \label{eq:position weights1} ,\\
\vo_i &=\sum_j a_{i, j} \vv_j\label{eq:token summation} , 
\end{align}
where $\boldsymbol{W}_Q$, $\boldsymbol{W}_K$ and $\boldsymbol{W}_V$ are query, key and value matrix $ \in R^{d \times d} $, respectively, and the attention weight of the $i$-th mixed token $\vo_i$ on the $j^{th}$ context token is $a_{i, j}$. 
According to Eq.~\ref{eq:Q}, Eq.~\ref{eq:K}, Eq.~\ref{eq:V}, we then have:
\begin{align}
\vq_i \boldsymbol{k}_j^{\top} &=\left(\vx_i+\vp_i\right) \boldsymbol{W}_Q \boldsymbol{W}_K^{\top}\left(\vx_j+\vp_j\right)^{\top} \notag
 \\ &= \left(\vx_i\right) \boldsymbol{W}_Q \boldsymbol{W}_K^{\top}\left(\vx_j\right)^{\top}+ \left(\vp_i\right) \boldsymbol{W}_Q \boldsymbol{W}_K^{\top}\left(\vx_j\right)^{\top} \notag \\ &+ \left(\vx_i\right) \boldsymbol{W}_Q \boldsymbol{W}_K^{\top}\left(\vp_j\right)^{\top}+ \left(\vp_i\right) \boldsymbol{W}_Q \boldsymbol{W}_K^{\top}\left(\vp_j\right)^{\top}. \label{eq:position weights2}
\end{align}

As shown in Eq.~\ref{eq:position weights1},  Eq.~\ref{eq:token summation}, Eq.~\ref{eq:position weights2}, the attention weight $a_{i,j}$ and the output mixed token $\vo_i$ are subject to the position embedding.
However, this mechanism is nonfunctional in Fourier neural operator mixer. The reason is that in Fourier neural operator mixers, attention weights are learned instead of obtaining from the dot product of the query vector for the $i$-th token $\vq_i$ and the key vector for th$j$-th token $\vk_j$ which includes terms related to position embeddings as shown in Eq.~\ref{eq:position weights2}.

\subsubsection{Fourier Neural Operators(FNO)}
\begin{figure}[t]
  \centering
  % \fbox{\rule{0pt}{0.5in} \rule{0.9\linewidth}{0pt}}
\includegraphics[width=1\linewidth]{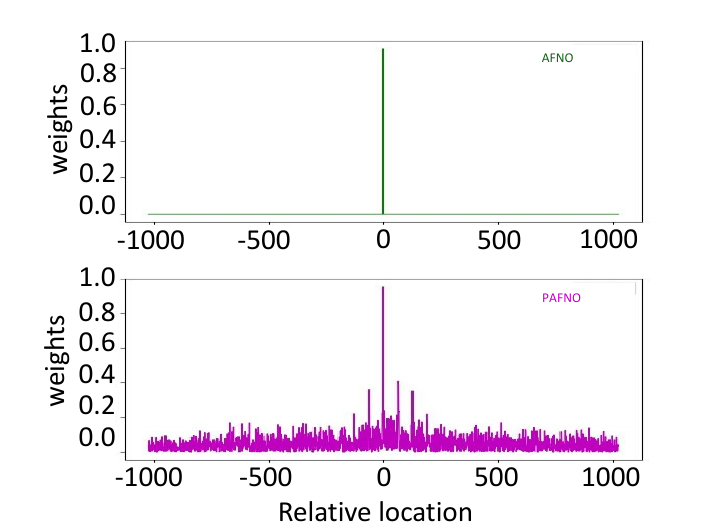}
   \caption{ 
   % \fontsize{0.5bp}{4bp} 
   \small Adaptive weights of PAFNO (bottom) to neighbours than AFNO (top). Weights (Y-axis) are from inverse DFT coefficients of $\lambda_n$ and X-axis denotes the spatial distance to the pivot token.}
   \label{fig:weights}
\end{figure}
To illustrate why the position embedding is nonfunctional in the FNO mixers, we compare 
the discrepancy between the dot-product self-attention mixer and the FNO mixer. 
In FNO, instead of using the self-attention mechanism to mix information among input tokens, discrete Fourier transform (DFT) is applied to decrease the computational complexity from $O(N^2)$ to $O(N\cdot \log N)$~\cite{li2020fourier}. The token mixing process can be formulated as follows:
 \begin{align}
    \vx^{\omega}_n\ &= {\rm DFT}(\{\vx\})_n \label{:temporal_afno_1},\\
    % \boldsymbol{W}^{\omega}_n\ &= {\rm DFT}(\{\tilde{K}^{\text{in}}\})_n \label{:temporal_afno_1},\\
    \tilde{\vx}^{\omega}_n &= \boldsymbol{W}^{\omega}_n \vx^{\omega}_n \label{:temporal_afno_2},\\
    \vx^{\text{out}}_n\ &= {\rm IDFT}(\{\tilde{\vx}^{\omega}\})_n.\label{:temporal_afno_3}
\end{align}
DFT/IDFT$(\{\cdot\})$ means performing DFT or IDFT on a sequence of signals. When comparing Eq.~\ref{:temporal_afno_1}, Eq.~\ref{:temporal_afno_2}, Eq.~\ref{:temporal_afno_3} with Eq.~\ref{eq:Q}, Eq.~\ref{eq:K}, Eq.~\ref{eq:V}, Eq.~\ref{eq:position weights1}, Eq.~\ref{eq:token summation}, the most significant difference between FNO and the dot-product self-attention is that FNO directly learns the token mixing matrix $\boldsymbol{W}^{\omega}_n$ in the frequency domain without considering position embedding as Eq.~\ref{eq:token summation}, which leads to the futility of absolute position embedding in FNO mixer.

Fortunately, as introduced in~\cite{guibas2021adaptive}, FNO can be considered as a global convolution, which naturally provides position-related weights similar to CNN. 
% $\tilde{K}_{0}$ is assigned to $\boldsymbol{x}_0$ when computing $\boldsymbol{o}_0$ and $\tilde{K}_{0}$ is assigned to $\boldsymbol{x}_i$ when computing $\boldsymbol{o}_i$.
However, FNO introduces $N\cdot D^2$ parameters, which can easily enlarge the complexity of the model ($N$ is the length of the token sequence). In order to simplify the model, adaptive Fourier neural operator (AFNO)~\cite{guibas2021adaptive} is proposed by reducing the number of frequency filters $\boldsymbol{W}^{\omega}_n$ from $N$ to $1$ (i.e., $\boldsymbol{W}^{\omega}$). But the limitation on the number of filters leaves only a unique filter to be used in the space domain. As a result, AFNO loses the position-related weights provided by different $\boldsymbol{W}^{\omega}_n$ in FNO, as shown in Figure~\ref{fig:weights}.

\subsubsection{PAFNO}
To address the issue mentioned above, we propose PAFNO to introduce positional information into the AFNO, which is visualized in Figure~\ref{fig:weights}. 

From Eq.~\ref{eq:V}, Eq.~\ref{eq:token summation} can be reformulated as:
\begin{align}
    \vo_i &=\sum_j a_{i, j} \vv_j \notag \\
                    &= \sum_j a_{i, j} \left(\vx_j+\vp_j\right) \boldsymbol{W}_V \notag \\
                    &= \sum_j \left(\vx_j+\vp_j\right)  \boldsymbol{K}_{i, j} \label{eq:self-attention-convolution}\\
    \boldsymbol{K}_{i, j} &= a_{i, j} \boldsymbol{W}_V \label{eq:k_ij}
    % \boldsymbol{o}_i &=\sum_j \left(\boldsymbol{x}_j+\boldsymbol{p}_j\right) \boldsymbol{K}_{i, j}. \label{eq:self-attention-convolution}
\end{align}
As shown in~\cite{li2020fourier}, it is assumed that $\boldsymbol{K}_{i, j}$ is a function of the distance between the pivot token $\boldsymbol{x}_i$ and the context token $\boldsymbol{x}_j$. Based on this assumption, $\boldsymbol{K}_{i, j}$ in Eq.~\ref{eq:self-attention-convolution} could be converted into a function of $i-j$ denoted as $\tilde{K}_{i-j}$, and Eq.~\ref{eq:k_ij} is reformulated as a convolution:
\begin{align}
    \vo_i &=\sum_j \left(\vo_j+\vp_j\right) \tilde{K}_{i-j} \label{eq:global_conv} \notag \\
          &= (\left(\vo+\vp\right) \ast \tilde{K})[i] .
\end{align}
According to Eq.~\ref{eq:k_ij}, an naive assumption for the possible form of the filter $\tilde{K}_{i-j}$ could be:
\begin{align}
\tilde{K}_{i-j} &= a_{i-j} \boldsymbol{W}_V .
\end{align}
DFT are then applied on $\tilde{K}$ to obtain the token mixing matrix $\boldsymbol{W}^{\omega}_n$ in the frequency domain as follows:
\begin{align}
    \boldsymbol{W}^{\omega}_n &= \sum_{k} \tilde{K}_k e^{-j \cdot \frac{2 \pi}{N} \cdot kn}, \notag \\
    &= (\sum_{k} a_k e^{-j \cdot \frac{2 \pi}{N} \cdot kn} )\boldsymbol{W}_V \label{eq:lambda},
\end{align}
where $k=i-j$. Based on Eq.~\ref{eq:lambda}, we then design a set of new frequency filters by introducing a set of learnable coefficients $\{\lambda_n\}$ into the AFNO as follows:
\begin{align}
    \boldsymbol{W}^{\omega}_n &= \lambda_{n}  \cdot \boldsymbol{W}^{\omega}\label{:pos-aware_filter},
\end{align}
where $\boldsymbol{W}^{\omega} \in R^{d\times d}$ is the unique frequency filter used in AFNO. Consequently, our PAFNO introduces the relative positional prior among the input tokens into the token mixer as FNO without significantly increasing the complexity of the model.

% Please refer to our supplementary materials for more discussion.

% Specifically, we denote the input tokens as $\{z^{\text{in}}_n\}_{n=1}^N$ and the MLP weights as $\mW^\omega$. Then the output of our PAFNO $\{z^{\text{out}}_n\}_{n=1}^N$ can be produced as follows:
% \begin{align}
%     \{z^{\omega}_n\} &= {\rm DFT}(\{z^{\text{in}}_n\}) \label{:temporal_afno_1},\\
%     \tilde{z}^{\omega}_n &=\lambda_n W^\omega z^{\omega}_n \label{:temporal_afno_2},\\
%     \{z^{\text{out}}_n\} &= {\rm IDFT}(\{\tilde{z}^{\omega}_n\}),
% \end{align}
% where ${\rm DFT}(\cdot)$ and ${\rm IDFT}(\cdot)$ are the discrete Fourier transform and inverse discrete Fourier transform, 
% % functions, 
% respectively. Thus, our PAFNO leverages the position information to mix token information in the frequency domain by assigning different coefficients for tokens at different positions.

\begin{table*}[t]
 \centering
 \resizebox{1.96 \columnwidth}{!}{
 \begin{tabular}{l|ccc|ccc}
 \toprule
    \multirow{2}{*}{Method}& \multicolumn{3}{|c}{RMSE(3/ 5 days)$\downarrow$}
     & \multicolumn{3}{|c}{ACC(3/ 5 days)$\uparrow$} \\
     & Z500($m^2s^{-2}$) & T850(K) & T2M(K) & Z500 & T850 & T2M \\
 \midrule
 Weekly climatology~\cite{rasp2020weatherbench} & 816 & 3.50 & 3.19 &0.65&0.77&0.85 \\
 \midrule
 T42~\cite{rasp2020weatherbench} & 489/743 & 3.09/3.83 & 3.21/3.69 & 0.90/0.78 & 0.86/0.78 & 0.87/0.83 \\
 T63~\cite{rasp2020weatherbench} & 268/463& 1.85/2.52 & 2.04/2.44 & 0.97/0.91 & 0.94/0.90 & 0.94/0.92 \\
 IFS~\cite{rasp2020weatherbench} & \textbf{154/334}& \textbf{1.36}/\textbf{2.03} & 1.35/1.77& \textbf{0.99}/\textbf{0.95} & \textbf{0.97}/\textbf{0.93} & \textbf{0.98}/\textbf{0.96} \\
 \midrule
 Na\"ive CNN~\cite{rasp2020weatherbench} & 626/757 & 2.87/3.37 &$\_$/$\_$ & 0.81/0.71 & 0.85/0.79& $\_$/$\_$\\
 Cubed UNet~\cite{weyn2020improving} & 373/611 & 1.98/2.87 &$\_$/$\_$ & 0.87/0.73 & $\_$/$\_$& 0.85/0.67 \\
 ResNet (pretrained)~\cite{rasp2021data} & 284/499 & 1.72/2.41 & 1.48/1.92 & 0.96/0.88 &0.95/\underline{0.90} &\underline{0.97}/\underline{0.95} \\
 FourCastNet~\cite{pathak2022fourcastnet} & 240/480& 1.50/2.50 & 1.50/2.00 &0.96/0.84& $\_$/$\_$ &0.88/0.75 \\
 SwinVRNN~\cite{hu2023swinvrnn} & 219/397& 1.47/\underline{2.06} & \textbf{1.25}/\textbf{1.66} & $\_$/$\_$ & $\_$/$\_$ & $\_$/$\_$ \\
%  SwinVRNN & $219$/$397$ & $1.47$/$2.06$ & $1.25$/$1.66$ &&& \\
 \midrule
WeatherFormer (ours) & \underline{181}/\underline{366} & \textbf{1.36}/2.07 & \underline{1.27}/\underline{1.72} &\underline{0.99}/\underline{0.96}&\underline{0.96}/\underline{0.90}& \underline{0.97}/0.94 \\
 \bottomrule
 % \multicolumn{7}{l}{\begin{tabular}{l}
 %      \emph{Note.} Latitude-weighted RMSE and ACC scores for 3 day, 5 day forecasts of Z500, T850 and T2M \\ are reported. For RMSE, the lower the better. For ACC, the higher the better. 
 % \end{tabular}}
 \end{tabular}
}
 % \vspace{-4mm}
  \caption{WeatherFormer v.s. state-of-the-art NWP methods on the WeatherBench dataset.}
\label{tab:nwp baselines}
\end{table*}

\subsection{Augmentations}
% 4.1 motivation of introducing weather data specified data augmentaion
% 4.2 earth rotation augmentation: motivation and how
% 4.3 error overlapping augmentation

% 4.1 motivation of introducing weather data specified data augmentaion
% The success of data augmentation for computer vision tasks has proven its effectiveness to ease the overfitting issue of deep neural networks. However, the widely-used image data augmentation strategies such as color transformation are not able to be adopted to the weather state data. 
\paragraph{ Rotation.}Earth rotation augmentation is proposed to utilize rotation equivariance in data.
As shown in Figure \ref{fig:point_mapping}, the weather state data is collected and predicted at the intersection of the lines of latitude and longitude, and then transferred to a weather state map. When the Earth rotates over the axis determined by the poles, the weather state map should rotate horizontally, which means the weather state map is rotation equivariance. Based on this observation, we introduced an earth rotation strategy, where the input weather states are horizontally rotated at a random distance in the training stage. By doing so, additional training samples are created to prevent our WeatherFormer from overfitting to the fixed grid of the training data and concentrate on the interior patterns over the context of weather states.

\paragraph{Noise.}
Noise augmentation is exploited to mitigate long-term error accumulation by roughly simulating the prediction error. Compared with methods requiring fine-tuning ~\cite{pathak2022fourcastnet}, noise augmentation half the energy demand without appreciable performance drop.

% Furthermore, it is observed that when we generate long-term predictions in an autoregressive manner, it can easily cause the error accumulation problem. Previous work 
% FourcastNet~\cite{pathak2022fourcastnet} attempted to address this problem by using a two-stage training strategy. Specifically, the FourcastNet is first pretrained to predict the weather states at the next time step, and then the pretrained model is finetuned by autogressively predicting the weather states at the next two time steps. However, the two-stage training strategy brings additional computational and time cost, which requires a large amount of computation resource. To this end, we introduce Error Overlapping augmentation to ease the error accumulation problem by simply adding random noise on top of the input weather states during the training stage. In this way, our WeatherFormer become more robust to prediction error without bringing any additional cost in the training stage, which can benefit long-term weather state prediction.

\section{Experimental Results}

This section first details the experimental setup in~\ref{sec:setup}, then compare our WeatherFormer with other state-of-the-art NWP methods on the Weatherbench dataset to verify the effectiveness of our WeatherFormer in~\ref{sec: weatherbench}. 
% In addition, our WeatherFormer can be generalized to the video prediction task, and we show the experimental results on the video prediction benchmarks Moving MINIST and TrafficBJ in \cref{sec:other dataset}.
Moreover, in order to investigate the contributions of our proposed components, ablation studies are conducted on the Weatherbench dataset and provide detailed analysis in~\ref{sec: ablation}. 
% Finally, the typhoon prediction generated by our WeatherFormer is visualized in supplementary.

\subsection{Experiments Setup}
\label{sec:setup}

\paragraph{Datasets.} We evaluate 
our WeatherFormer on the numerical weather prediction benchmark dataset, \textbf{Weatherbench}~\cite{rasp2020weatherbench}. 
It is a medium-range weather forecasting (specifically 3-5 days) dataset 
whose data is collected by downsampling from the ERA5 reanalysis data~\cite{hersbach2020era5} from 1979 to 2018. 
Weather states with three different scales are provided in Weatherbench, whose data points are sampled from the latitude and longitude with an interval of 5.625$^\circ$(32 $\times$ 64 grids), 2.8125$^\circ$(64 $\times$ 128 grids), and 1.40525$^\circ$(128 $\times$ 256 grids), respectively. 
In each grid, there are 13 vertical layers that include different weather states such as wind speed, geopotential, humidity, temperature, etc. The 32 $\times$ 64 grid weather state data are chosen during both the training and testing stages.

% To validate the effectivenss and generalization ability of our WeatherFormer, we also conduct experiments on two video prediction benchmark datasets, TrafficBJ~\cite{srivastava2015unsupervised} and MovingMNIST~\cite{srivastava2015unsupervised}. \textbf{TrafficBJ} contains the trajectory data in Beijing collected from taxicab GPS with two channels, i.e. inflow or outflow defined in~\cite{zhang2017deep}. \textbf{MovingMNIST} contains 10000 samples in both the training and testing set, and it consists of two digits independently moving within the 64 × 64 grid and bouncing off the boundary. 
% We follow ~\cite{gao2022earthformer} to pre-generate the digits to construct a fixed dataset.

\paragraph{Implementation Details}
%数据细节
On Weatherbench, we apply the $32\times64$ data with 6-hour time intervals, in which each data point contains weather states and 2 constant variables (i.e., land binary mask and orography). For the perdition results, the 69 dynamic weather states are forecast. In the training stage, the proposed WeatherFormer predicts weather states at the next 6 hours by using the weather states of the previous 36 hours as the input (i.e., a sequence of 6 weather state maps with a temporal interval of 6 hours). 
%模型细节
The patch size used to tokenize the weather state input is selected to be 1 in the spatial dimension and 2 in the temporal dimension, and the embedding dimension of the token is 1024. The number of layers of SF-Blocks is 12 and other settings are the same as in~\cite{pathak2022fourcastnet}. 
% {\color{gjc} things about data augmentation}
Moreover, we horizontally rotate the input data by a random distance range from 0 to 64 with a 50\% probability for the Earth Rotation augmentation, and a normal Gaussian noise with a variance of 0.1 is used for noise  augmentation.
%训练细节. 
 AdamW is our optimizer with a learning rate of 0.0005 in a cosine learning rate scheduler. Our WeatherFormer is trained for 80 epochs with the first 5 epochs for warmup. The training needs 8 A100 GPUs with a mini-batch size of 4 in each gpu for 62 hours.   
 \paragraph{Evaluation Metrics} As used in \cite{pathak2022fourcastnet,hu2023swinvrnn}, we also use the weighted Root Mean Square Error (RMSE) and Accuracy (ACC) as the metrics for evaluation, where smaller RMSE and bigger ACC means better performance.

\begin{table}[t]
\centering
% \resizebox{\textwidth}{!}{
\resizebox{\columnwidth}{!}{
\begin{tabular}{c c c c| c c}
\toprule
% & & & \multicolumn{4}{c}{RMSE (1 days/ 3 days/ 5 days)} \\
% \cmidrule{4-7}
SF-B & PAFNO & ER  & noise &  RMSE &  ACC \\
\midrule
 &  &  &  & $120/344/618$ & $0.99/0.95/0.84$   \\
\checkmark & & & & $106/286/511$ & $0.99/0.96/0.87$   \\
\checkmark & \checkmark & &  & $98/266/486$ & $0.99/0.96/0.88$ \\
\checkmark & \checkmark & \checkmark &  & $93/256/473$ & $0.99/0.97/0.89$ \\
\checkmark & \checkmark & \checkmark & \checkmark  & $89/241/444$ & $0.99/0.97/0.90$  \\
\bottomrule
% \multicolumn{6}{l}{\begin{tabular}{l}
% \emph{Note.} We use WeatherFormer-B to report ablation study results. ER denotes earth rotation and EO means error overlapping.
% \end{tabular}} \\
\end{tabular}
}
\caption{Ablation Study for the Designed Components of WeatherFormer on WeatherBench Dataset. Our baseline is AFNO. SF-B represents SF-blocks, ER denotes earth rotation augmentation. 1/3/5 days Z500's metrics are displayed.}
\label{tab:ablation}
\end{table}

\subsection{Results on the Weatherbench dataset} 
\label{sec: weatherbench}
On Weatherbench, our WeatherFormer predicts 69 weather states at every data point position. To verify the effectiveness of our WeatherFormer, we compare the prediction performance of our WeatherFormer with other state-of-the-art NWP methods in terms of 3 typical weather states: 1) the geopotential at the height of atmospheric pressure of 500hPa (i.e., Z500); 2) the temperature at the height of atmospheric pressure of 850hPa (i.e., T850); and 3) the temperature at the height of 2 meters (i.e., T2M). 

We report RMSE and ACC of our WeatherFormer in terms of Z500, T850, and T2M in Table~\ref{tab:nwp baselines}. Among the compared state-of-the-art NWP methods, T42, T63, and IFS are the traditional physics based NWP methods, while Na\"ive CNN, Cubed UNet, ResNet (pretrained),  FourCastNet, and SwinVRNN use deep neural networks to predict weather states in a data-driven manner. 
% It is interesting to note that all these compared data-driven NWP methods only consider the spatial information among the input weather states, while our WeatherFormer can model the spatial-temporal relationship to predict future weather states. 
As shown in Table~\ref{tab:nwp baselines}, for IFS, WeatherFormer performs better than its lower-resolution version, e.g., T42 (64 $\times$ 128) and T63 (90 $\times$ 180), while comparing lower-resolution WeatherFormer to high-resolution IFS is unfair.  IFS is an ensemble of 51 physical models that use a higher resolution (e.g., $720 \times 1440$) and more inputs (hourly) than ours (one model, $32 \times 64$, six-hourly). Despite this, we still perform better in T2M, comparable in T850 and Z500. 
Moreover, the energy consumed by IFS running a year is about 150000 times the consumption of training a WeatherFormer ~\cite{bauer2020ecmwf}.
When compared with SwinVRNN, the synthesis performance of our WeatherFormer is better, and the training consumption of WeatherFormer (80 epochs and 1-step training) is markedly lower than SwinVRNN (200 epochs and multistep RNN training). 

% our WeatherFormer significantly outperforms any other state-of-the-art data-driven NWP methods at all three weather states in terms of all evaluation metrics, which demonstrates the effectiveness of the proposed space-time transformer. 

% It is observed that despite the data-driven NWP method can bring a large amount of reduction in the computational cost, their weather prediction performance is always inferior to that of the physics based NWP methods by a large margin. However, our WeatherFormer can narrow down the performance gap between the data-driven and the physics based NWP methods, and on T2M, our model even outperforms the best physics based NWP method (i.e., IFS) in terms of RMSE. Note that, all the observations mentioned above also can be found in the other 66 weather states.

% \begin{figure*}[t]
%   \centering      \includegraphics[width=0.98\textwidth]{fig/fig4_trackNew.jpg}
%     \caption{Wind speed at 10m above the surface.  The redder the color is, the higher the wind speed is. Red box identifies the location of Rumbia tropical cyclone at a given timestamp. }
%     \label{fig:Rumbia}
%     \vspace{-2mm}
% \end{figure*}

\begin{table}[t]
 \centering
 \resizebox{\columnwidth}{!}{
 \begin{tabular}{c c|ccc}
 \toprule
    \multirow{2}{*}{TS}&\multirow{2}{*}{noise} &\multicolumn{3}{|c}{RMSE(1/ 3/ 5 days)} \\
     &  &Z500($m^2s^{-2}$) & T850(K) & T2M(K) \\
\midrule
% noise & 100/276/512& 1.20/1.77/2.58 & 1.30/1.68/2.18 &0.995/0.963/0.877 & 0.970/0.937/0.867 &0.968/0.947/0.912 \\
% noise with ft & 96/261/492 & 1.19/1.75/2.55 & 1.29/1.69/2.18 & 0.995/0.966/0.883 & 0.971/0.938/0.869 &0.968/0.946/0.912 \\
% w/o noise & 102/286/536 & 1.20/1.84/2.71 & 1.31/1.77/2.32 & 0.996/0.971/0.899 & 0.975/0.947/0.886 &0.976/0.959/0.930 \\
% w/o noise with ft & 99/277/521 & 1.19/1.80/2.64 & 1.30/1.75/2.27 & 0.995/0.963/0.874 & 0.971/0.934/0.863 &0.968/0.943/0.908 \\
& & 102/286/536 & 1.20/1.84/2.71 & 1.31/1.77/2.32 \\
\checkmark & & 99/277/521 & 1.19/1.80/2.64 & 1.30/1.75/2.27 \\
& \checkmark & 100/276/512& 1.20/1.77/2.58 & 1.30/1.68/2.18  \\
\checkmark & \checkmark & 96/261/492 & 1.19/1.75/2.55 & 1.29/1.69/2.18 \\
 % & 86/232/433 & 1.08/1.59/2.30 & 1.10/1.45/1.90 &0.995/0.961/0.868 & 0.970/0.932/0.857 &0.968/0.942/0.903 \\
% without noise with ft & 89/241/445& 1.09/1.60/2.34 & 1.11/1.45/1.90 &0.996/0.970/0.897 & 0.975/0.946/0.885 &0.976/0.960/0.931 \\
\bottomrule
 % \multicolumn{7}{l}{\begin{tabular}{l}
 %      \emph{Note.} Latitude-weighted RMSE and ACC scores for 3 day, 5 day forecasts of Z500, T850 and T2M \\ are reported. For RMSE, the lower the better. For ACC, the higher the better. 
 % \end{tabular}}
 \end{tabular}
 }
  \caption{RMSE of Z500, T850 and T2M predicted by FourcastNet by using different training strategies on the Weatherbench dataset. TS denotes two-stage training strategy.}
  \label{tab:noise}
 \end{table}

% \subsection{Results on Other Datasets.} \label{sec:other dataset}

\subsection{Ablation Study} \label{sec: ablation}
To investigate the effectiveness of the proposed components, we conduct an ablation study on the Weatherbench dataset. The experimental results are reported in Table~\ref{tab:ablation}.

\paragraph{Effectiveness of components.}
For the ablation study, we first design a baseline method by removing the temporal mixer from our WeatherFormer and using the AFNO as the token mixer. In order to accelerate the ablation study experiments, we also increase the patch size to reduce the complexity of the baseline method. We take the RMSE of the Z500 prediction at the next day as the example to analyze the prediction contribution of each proposed component. As shown in Table~\ref{tab:ablation}, our baseline method can achieve RMSE of 120, which is improved by adding temporal mixer into the SF-Block in our WeatherFormer. This suggests that the temporal trend of the past weather states is critical in the NWP task. Additionally, in our proposed PAFNO, we introduce position-related coefficients to mix token information by considering the impact of the token positions. When we replace the AFNO with the proposed PAFNO, the RMSE is further decreased from 106 to 98, which demonstrates the effectiveness of introducing the position-related coefficients. It is well-known that data-driven NWP methods often suffer from the training data overfitting problem. 
% {\color{gjc} data augmentation}
We adopt the Earth Rotation augmentation strategy during the training stage and observe that after applying the Earth Rotation augmentation during the training stage, another decrease in the RMSE (from 98 to 93) appeared on Z500. The results show that our Earth Rotation augmentation strategy is able to ease the overfitting issue to the training dataset for our WeatherFormer. Finally, by using the noise augmentation strategy, the RMSE further boosts to 89, indicating noise augmentation strategy reduces error accumulation.

\paragraph{Noise Augmentation v.s. Two-stage Training.} Error accumulation often happens on long-term prediction in a progressive manner. FourcastNet~\cite{pathak2022fourcastnet} attempted to ease this problem by using a two-stage training strategy, where the NWP model is first pretrained, and then fine-tuned to predict the weather states at the next two time steps in an autoregressive manner. To compare the effectiveness of our Error Overlapping augmentation and two-stage training strategy, we use FourcastNet as the baseline method, and report the RMSE of the Z500, T850 and T2M prediction at the next first, third, and fifth days by using different strategies in Table~\ref{tab:noise}. We can observe that despite the two-stage training strategy is able to improve the RMSE results, by only using the noise augmentation, FourcastNet can achieve better RMSE results at long-term prediction at all three weather states (see RMSE on the third and fifth days for second and third rows of Table~\ref{tab:noise}). This suggests that the noise augmentation can relieve the error accumulation issue better than the two-stage training strategy does. Additionally, for the two-stage training strategy, since a fine-tune training process is introduced, two times of training time is required which doubles the energy consumption. Moreover, in the fine-tune training process, as the NWP framework is trained with an autoregressive manner, a huge amount of memory and computation resources is needed. In contrast, noise augmentation can reduce the effect caused by error accumulation and improve long-term prediction performance without bringing any additional cost. Also, it is interesting to note that further gain can be observed when we apply noise augmentation with a two-stage training strategy, which demonstrates these two strategies are complementary.

% \subsection{Qualitative Results} \label{sec:qualtative}
% To further verify the effectiveness of our WeatherFormer, we visualize the prediction results of wind speed at 10m height above the surface generated by our WeatherFormer in Figure~\ref{fig:Rumbia} to exhibit its ability to predict extreme weather. As shown in Figure~\ref{fig:Rumbia}, the red boxes in the GT figures demonstrate that Typhoon Rumbia was formed as a tropical depression around August 15, 2018, on the Pacific Ocean, and after two days of moving, it made landfall in Shanghai on August 17, 2018. Our WeatherFormer successfully predicts the formation and the moving track of Rumbia (see the red boxes in the Pred figures in Figure~\ref{fig:Rumbia}). The paths in the ground truth and paths generated by  WeatherFormer are well aligned both in space and time. 

\section{Conclusion}
In this paper, we propose a space-time transformer-based network that perceives spatial-temporal information with the FFT-based mixer for weather forecasting. Specifically, we propose the PAFNO to maintain the parameter amount while making spatial mixer relative position-aware. Moreover, we notice the discreteness and rotation symmetry in weather data and introduce earth rotation augmentation. WeatherFormer achieves SOTA among data-driven methods on Weatherbench. On the one hand, this work demonstrates that data-driven NWP methods have tremendous potential to be applied in future weather prediction system. On the other hand, it is apparent that the challenges towards practice application are stout, and we anticipate more research interests are devoted into this newly emerging direction.

% and it also attains impressive performance on TrafficBJ and MovingMNIST. 

% \begin{figure*}[h!tbp]
%   \centering \includegraphics[width=0.98\textwidth]{fig/fig4_trackNew.jpg}
%     \caption{Wind speed at 10m above the surface. The more redder color represents the higher wind speed. \textcolor{red}{Red} box identifies the location of Rumbia tropical cyclone at a given timestamp. }
%     \label{fig:Rumbia}
% \end{figure*}

\newpage
% \pagebreak
\appendix
\section{Appendix}

\begin{figure*}[h!tbp]
  \centering \includegraphics[width=0.98\textwidth]{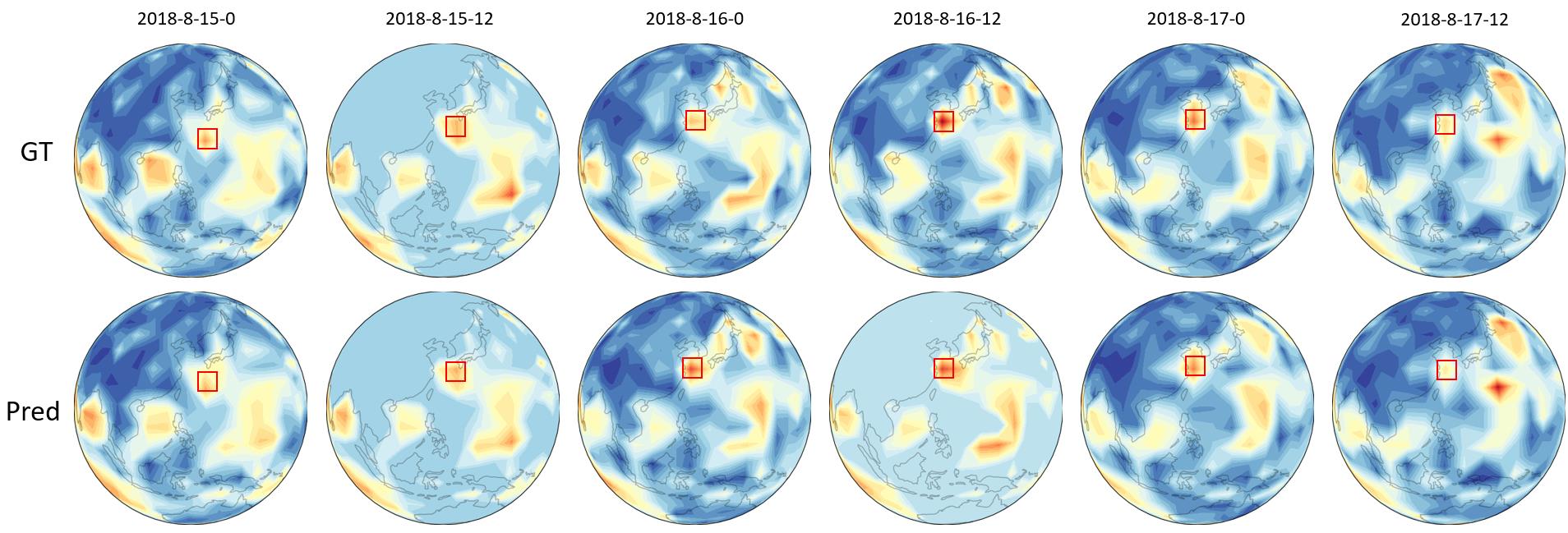}
    \caption{Wind speed at 10m above the surface. The more redder color represents the higher wind speed. \textcolor{red}{Red} box identifies the location of Rumbia tropical cyclone at a given timestamp. }
    \label{fig:Rumbia}
\end{figure*}

% The file named.bst is a bibliography style file for BibTeX 0.99c

\subsection{Qualitative Results} \label{sec:qualtative}

To further verify the effectiveness of our WeatherFormer, we visualize the prediction results of wind speed at 10m height above the surface generated by our WeatherFormer in Figure~\ref{fig:Rumbia} to exhibit its ability to predict extreme weather. As shown in Figure~\ref{fig:Rumbia}, the red boxes in the GT figures demonstrate that Typhoon Rumbia was formed as a tropical depression around August 15, 2018, in the Pacific Ocean, and after two days of moving, it made landfall in Shanghai on August 17, 2018. Our WeatherFormer successfully predicts the formation and the moving track of Rumbia (see the red boxes in the Pred figures in Figure~\ref{fig:Rumbia}). The paths in the ground truth and paths generated by  WeatherFormer are well aligned both in space and time. 
\subsection{Evaluation Metrics} Following~\cite{rasp2020weatherbench}, we apply latitude-weighted root-mean-square error (RMSE) and anomaly correlation coefficient(ACC) to evaluate our WeatherFormer for the NWP task. The latitude-weighted RMSE is calculated as follows:
\begin{align}
L(j) &= \frac{\cos \operatorname{lat}(j)}{\frac{1}{N_{l a t}} \sum_j^{N_{l a t}} \cos \operatorname{lat}(j)}, \\
MSE &=  \frac{1}{N_{\text {lat }} N_{\text {lon }}} \sum_j^{N_{\text {lat }}} \sum_k^{N_{\text {lon }}} L(j)\left(y_{ j, k}-\hat{y}_{ j, k}\right)^2, \\
RMSE &= \frac{1}{N_{\text {sample }}} \sum^{N_{\text {sample }}} \sqrt{MSE}.
\end{align}
$N_{\text {sample }}$ is the number of samples. $N_{\text {lon }}$ and $N_{\text {lat}}$ are the number of grid points along longitude and latitude, respectively. $y_{j,k}$ and $\hat{y}_{j,k}$ indicate the predicted weather states and ground truth weather states at the $j$-th and $k$-th data point along the latitude and longitude, respectively. 
ACC is calculated as follows:
 \begin{gather}
     c_{j,k} = \frac{1}{N_{time}}\sum{\hat{y}_{j,k}}, \\
y_{i, j, k}^{\prime} = y_{i, j, k} - c_{j,k},\\
\hat{y}_{i, j, k}^{\prime} = \hat{y}_{i, j, k} - c_{j,k}, \\
\mathrm{ACC} =\frac{\sum_{i, j, k} L(j) y_{i, j, k}^{\prime} \hat{y}_{i, j, k}^{\prime}}{\sqrt{\sum_{i, j, k} L(j) y_{i, j, 
k}^{\prime 2} \sum_{i, j, k} L(j) \hat{y}_{i, j, k}^{\prime 2}}}.
 \end{gather}
$N_{time}$ denotes the number of samples in the training set. Lower RSME indicates better prediction performance, while higher ACC indicates better prediction performance.

\bibliographystyle{named}
\bibliography{ijcai23}

\end{document}